\UseRawInputEncoding

\documentclass[a4paper, 10pt, conference]{ieeeconf}      

\IEEEoverridecommandlockouts                              

\title{\LARGE \bf
An End-to-end Deep Reinforcement Learning Approach for the Long-term Short-term Planning on the Frenet Space}

\usepackage{cite}
\usepackage{amsmath,amssymb,amsfonts}
\usepackage{algorithmic}
\usepackage{graphicx}
\usepackage{textcomp}
\usepackage{xcolor}
\usepackage{subfig}
\usepackage{float}
\usepackage{multirow}
\usepackage{tikz}
\usepackage{hyperref}

\usepackage{amsthm}

\newcommand{\DrawBlackPercentageBar}[1]{%
  \begin{tikzpicture}
    \fill[color=black]   (0.0 , 0.0) rectangle (#1*8ex , 1.5ex );
    \fill[color=gray] (#1*8ex  , 0.0) rectangle (8.0ex, 1.5ex);
  \end{tikzpicture}%
}

\newcommand{\DrawRedPercentageBar}[1]{%
  \begin{tikzpicture}
    \fill[color=red]   (0.0 , 0.0) rectangle (#1*8ex , 1.5ex );
    \fill[color=gray] (#1*8ex  , 0.0) rectangle (8.0ex, 1.5ex);
  \end{tikzpicture}%
}

\def\BibTeX{{\rm B\kern-.05em{\sc i\kern-.025em b}\kern-.08em
    T\kern-.1667em\lower.7ex\hbox{E}\kern-.125emX}}

\author{Majid Moghadam$^{1}$, Ali Alizadeh$^{2}$, Engin Tekin$^{1}$, and Gabriel Hugh Elkaim$^{1}$ 
\thanks{$^{1}$Majid Moghadam, Engin Tekin, and Gabriel Hugh Elkaim are with Faculty of Electrical and computer Engineering, University of California Santa Cruz (UCSC), California, U.S
        {\tt\small \{mamoghad, etekin, elkaim\}@ucsc.edu}}%
\thanks{$^{2}$Ali Alizadeh is with the Department of Mechatronics Engineering, Istanbul Technical University (ITU), alizadeha@itu.edu.tr
        {\tt\small alizadeha@itu.edu.tr}}%
\thanks{This work has been submitted to International Conference on Robotics and Automation (ICRA 2021) for review.}
}

\begin{document}

\maketitle
\thispagestyle{empty}
\pagestyle{empty}
\theoremstyle{remark}
\newtheorem*{remark}{\textbf{Remark}}

\begin{abstract}
Tactical decision making and strategic motion planning for autonomous highway driving are challenging due to the complication of predicting other road users' behaviors, diversity of environments, and complexity of the traffic interactions. This paper presents a novel end-to-end continuous deep reinforcement learning approach towards autonomous cars' decision-making and motion planning. For the first time, we define both states and action spaces on the Frenet space to make the driving behavior less variant to the road curvatures than the surrounding actors' dynamics and traffic interactions. The agent receives time-series data of past trajectories of the surrounding vehicles and applies convolutional neural networks along the time channels to extract features in the backbone. The algorithm generates continuous spatiotemporal trajectories on the Frenet frame for the feedback controller to track. Extensive high-fidelity highway simulations on CARLA show the superiority of the presented approach compared with commonly used baselines and discrete reinforcement learning on various traffic scenarios. Furthermore, the proposed method's advantage is confirmed with a more comprehensive performance evaluation against 1000 randomly generated test scenarios.
\\
\\
Code: \textnormal{https://github.com/MajidMoghadam2006/RL-frenet-trajectory-planning-in-CARLA}
\end{abstract}

\section{INTRODUCTION}

Driving in highway traffic is a challenging task, even for a human, that requires intelligent decision-making for long-term goals and cautious short-term trajectory planning to execute decisions safely. Several methods have been proposed for the decision-making of autonomous vehicles on a highway driving task. Most of the studies approached as a control problem \cite{taylor1999comparative, hatipoglu2003automated}. Recently, reinforcement learning (RL) approaches have presented a decent alternative to the optimal lane-changing problem \cite{hoel2018automated, alizadeh2019automated, yavas2020new, liu2020decision}. However, the main challenge is to translate these decisions to safe trajectories.

Trajectory planning, has been addressed by multiple studies. Claussmann et al. \cite{claussmann2019review} distinguish the space configuration for the path planning into three main categories: i.e., sampling points \cite{li2014unified}, connected cells \cite{yu2016semantic}, and lattice representation \cite{werling2010optimal}. While each method has pros and cons, lattice enables predictive planning based on the moving obstacles surrounding ego while considering the kinematic constraints and vehicle motion primitives. In this work, we have utilized lattice representation on the Frenet frame to explore the space and generate the optimal trajectory. 
\begin{figure}
    \centering
    \includegraphics[width=\linewidth]{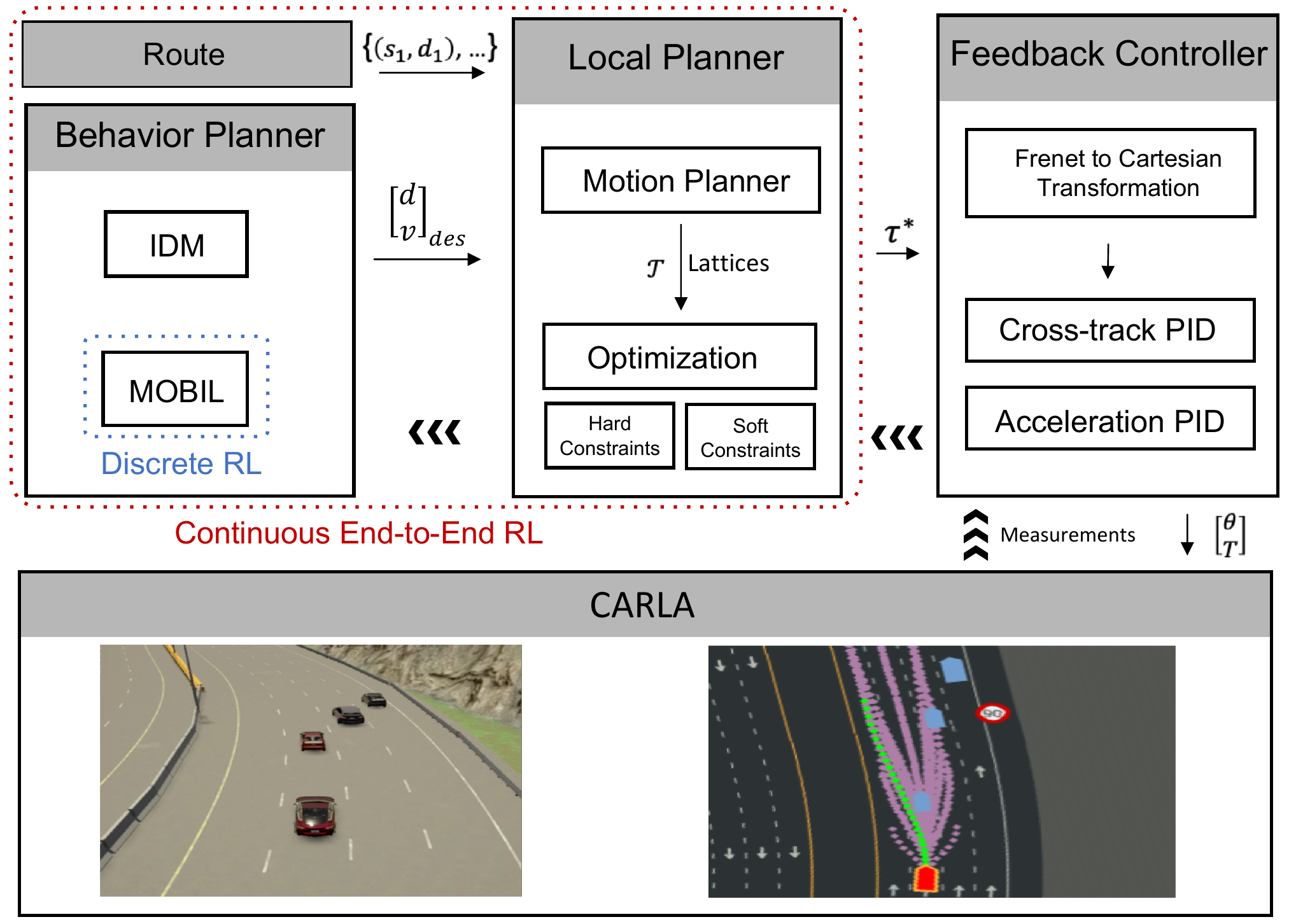}
    \caption{The proposed hierarchical architecture for the long-term decision-making and short-term trajectory planning in the Frenet space.}
    \label{fig:ADAS hierarchical arch.}
\end{figure}

Deep reinforcement learning (DRL) is an emerging and promising method to apply to autonomous driving problems. Several studies leverage the raw sensory measurements to generate steering angle and throttle values in an end-to-end manner \cite{yu2018intelligent, sallab2016end, sallab2017deep, aradi2018policy, nageshrao2019autonomous, jaritz2018end}. Although these approaches have shown convincing results in game environments, they suffer the steering and throttle fluctuations failing to provide smooth driving. DRL's has shown a satisfactory performance on behavior planning. Nevertheless, as pointed out in the recent survey paper \cite{aradi2020survey}, few studies applied DRL on motion planning for autonomous driving systems. Feher et al. \cite{feher2019hybrid} trained a DDPG agent to generate the waypoints for the vehicle to track. Initially, the planner generates a list of waypoints up to a fixed time horizon and step-size. Then, DRL modifies the lateral displacement of the waypoints to generate an optimal trajectory. This algorithm's main drawback is that it only focuses on lateral planning and fixes the longitudinal path, which provides a sub-optimal trajectory that may not be a suitable choice for complex driving environments like highway. Besides, the study did not provide a suitable performance measurement for the unseen scenarios in a simulation. Also, few papers deviate from vehicle motion restrictions and generate actions by stepping in a grid, like in classic decision-making \cite{nageshrao2019autonomous}.

This work introduces a novel approach toward motion planning using DRL, which fills the literature gap between decision-making and end-to-end DRL. The highlight of our contributions can be summarized as follows:
\begin{itemize}
\item We present a novel end-to-end continuous RL trajectory planning that explores the driving corridors \cite{bender2015combinatorial} in moving obstacles and generates spatiotemporal trajectories that safely navigate through traffic.
\item State representations for RL are in the Frenet frame \cite{werling2010optimal} that is for the first time in the literature to the best of our knowledge.
\item The presented agent utilizes continuous states and generates time-polynomial continuous trajectories. In contrast to most of the literature methods, we do not perform discretization anywhere in the process.
\item We utilize CARLA as a high-fidelity simulation environment to generate realistic highway traffic scenarios and exhibit the RL agents' performance and compare the RL agents with baseline approaches in qualitative and quantitative metrics.
\end{itemize}
\section{Planning on Frenet Space}
Although the literature for vehicle motion planning has many variations \cite{claussmann2019review, lopez2019new, li2015real, li2015practical, ziegler2014trajectory, ziegler2014making, jordan2019real}, most of the algorithms emulate a hierarchical architecture of behavior planning and trajectory generation. To have a valid baseline that represents the level of performance in the literature, we used state-of-the-art methods for each layer, such as Intelligent Driving Model (IDM) \cite{treiber2000congested} and Minimizing-Overall-Braking Induced-by-Lane-changes (MOBIL) \cite{kesting2007general} algorithms for behavior planner (BP) and a global optimization for trajectory generation. The overall architecture of the presented baseline is demonstrated in Fig. \ref{fig:ADAS hierarchical arch.}.

Behavior planner utilizes the 2D/3D environment model and commands high-level driving actions to safely maneuver through the traffic and maintain the target speed as in \cite{alizadeh2019automated}. BP consists of two components; the IDM as cruise control to maintain a safe distance with the leading vehicle and MOBIL that utilizes the ego's relative positions and velocities with the surrounding vehicles to generate safe lane change decisions. Manipulating IDM and MOBIL parameters enables us to achieve the various driving styles, i.e. agile and safe drivers. 

\begin{figure}
    \centering
    \includegraphics[width=\linewidth]{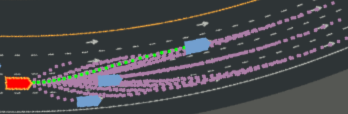}
    \caption{Lattice representation of {\color{purple}candidate polynomial trajectories} and {\color{green}optimal trajectory}}
    \label{fig:lattices}
\end{figure}

Local planner (LP) translates the BP's long-term decisions (target lane and speed) to optimal trajectories in a variable time-horizon window. LP operates in the Frenet frame \cite{werling2010optimal} to make the driving behavior less variant to the road curvatures than the surrounding actors' dynamics and interactions. On the Frenet space, the vehicle's position is specified in terms of longitudinal displacement along the road's arc ($s$) and the lateral offset from the highway shoulder ($d$), i.e. a vehicle driving on a lane with a fixed speed would have constant values for $\dot s$ and $d$. A trajectory $\tau$ on the Frenet space, can be represented by two time-polynomials $[s(t), d(t)]$. At each time-step, motion planner (MP) generates a set of candidate polynomial trajectories  $\mathcal{T}=\{\tau_1, ..., \tau_m\}$ known as lattices (Fig. \ref{fig:lattices}). Lattices are being generated by varying the trajectories' terminal manifolds \cite{werling2012optimal}: speed $v_f$, lateral position $d_f$, and arrival time $t_f$ respectively. The generated trajectories laterally connect to the optimal trajectory from the previous iteration and align with the road's arc. 

Using a simple linear search, we pick the optimal trajectory $\tau^*$ from the set $\mathcal{T}$. The algorithm eliminates infeasible trajectories that violate hard constraints, such as min/max speed, max acceleration, and potential collision with moving obstacles. Soft constraints in the optimization penalize an objective function in terms of safety, reliability, and comfort to select the optimal path:
\begin{equation}
    \mathbf{J}(\tau) = w_{o} J_{o} + w_{v} J_{v} + w_{a} J_a + w_j  J_{j} + w_{\dot \psi} J_{\dot \psi}
\label{e: objective function}
\end{equation}
where, $w_x$ weights control the trade-off between agility, safety, and comfort. $J_o$ minimizes the vehicle's lateral offset to the target lane center. $J_v$ includes the vehicle speed error in the cost function to drive at the desired speed commanded by BP. Finally, $J_a$, $J_j$, and $J_{\dot \psi}$ suppress the vehicle acceleration, jerk, and yawing rate to improve safety and comfort. The outcome of this process in LP is the optimal trajectory $\tau^*$ on the Frenet space. 

Finally, a feedback controller transforms $\tau^*$ from Frenet to Cartesian coordinate \cite{werling2010optimal} and utilizes separate lateral and longitudinal controllers to minimize cross-track and acceleration errors.

\section{DISCRETE RL FOR DECISION MAKING}
It is possible to replace the MOBIL with a discrete RL that receives the raw sensory measurements (ground truth in our simulations) and generates high-level actions. Discrete actions include lane change commands. We have the option of removing IDM and adding the accelerate/decelerate commands for the action list; however, we will show that IDM exhibits a decent performance in generating continuous value for the target speed. Thus, there is no need to discretize the longitudinal state. The action space for this agent consists of LefLaneCHange, RightLaneChange, and StayOnTheLane actions. 

The observation space and the neural network architecture are the same as end-to-end RL presented in the next section. We utilized the double deep Q-network (DQN) \cite{mnih2015human} with the experience replay memory for policy optimization, a reasonably stable discrete RL agent available in the literature.

\section{END-TO-END CONTINUOUS RL FOR DECISION MAKING AND MOTION PLANNING}
Formulating the motion planning problem in the reinforcement learning setting requires a careful understanding of the decision-making and trajectory generation hierarchy. In the experiments section we will show that the provided baseline framework on the Frenet space exhibits an adequate performance in generating feasible and safe trajectories. However, the degree of the optimality of the generated path inadvertently depends on the information that we tailor and incorporate in the objective function. Besides, discretizing lattices to find the best trajectory adds an extra level of inaccuracy to the solution's optimality. Alternatively, we can devise (an) RL agents that leverage deep neural networks to extract useful features from raw sensory measurements (ground truth in our experiments) and provide optimal trajectory in the continuous action space. One method is to adopt baseline architecture (see Fig. \ref{fig:ADAS hierarchical arch.}) but use two separate RL agents for BP and LP layers; a discrete RL for BP and a continuous RL for LP. In this case, incorporating the discrete RL's actions in the continuous RL's observation space would complicate the architecture since the feature extraction of the latter should be manipulated to compute observations with various natures separately. Besides, serially connecting RL agents would intensify the computations and destabilize the training process since both agents require to tune the exploration-exploitation trade-off simultaneously. An end-to-end agent can be a reasonable alternative where the agent receives raw sensory measurements and optimizes a policy that generates a continuous polynomial trajectory in the Frenet space. The term end-to-end, in contrast with the literature, does not imply utilizing raw measurements to generate actuation ($\theta, T$), but instead, to produce a continuous trajectory ($\tau^*$). Thus, the RL formulation problem boils down to three steps: action space determination, neural network architecture design, and the policy optimization.
\newline
\subsubsection*{Action Space}
\begin{figure}[!h]
    \centering
    \includegraphics[width=\linewidth]{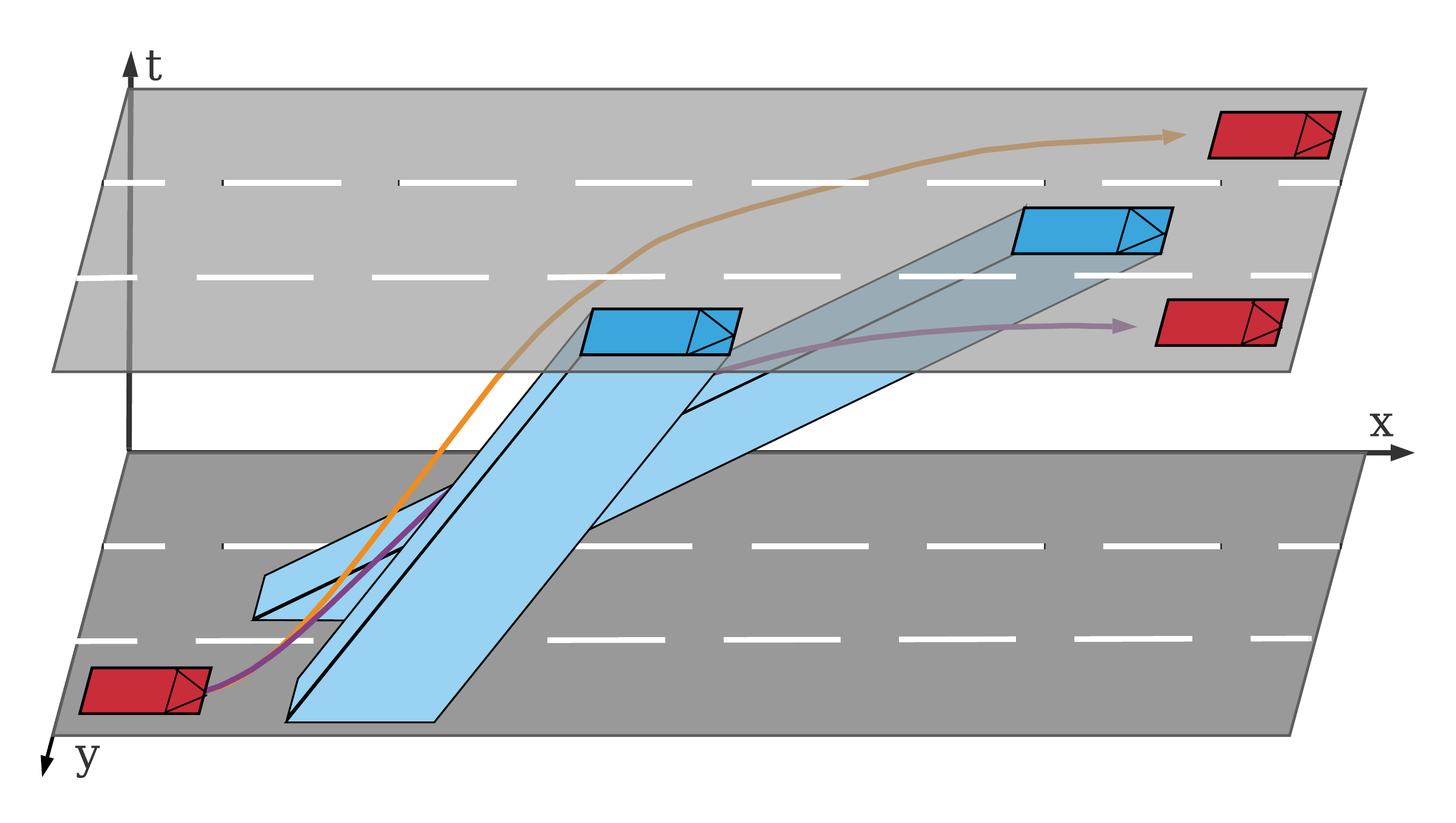}
    \caption{Spatiotemporal trajectories for moving obstacles and candidate paths visualized in driving corridors}
    \label{f:driving corridor}
\end{figure}

Instead of discretizing the terminal manifolds like Baseline approach, we plan to introduce an intelligent agent that generates a continuous spline in the Frenet space at each time-step. A polynomial trajectory, aka a lattice, can be characterized using three continuous values: $v_f$, $d_f$, and $t_f$. Each value has an acceptable range for the action sampling. Mapping these ranges into $[-1, 1]$ enables us to define the RL continuous action space as 
\begin{equation}
    \mathcal{A} = \{v_f, d_f, t_f\}
\end{equation}
where, each value is in the $[-1, 1]$ range. Exploring various regions in the action space is equivalent to examining different splines in the driving corridors \cite{bender2015combinatorial}, as illustrated in Fig. \ref{f:driving corridor}.
\newline
\subsubsection*{State Space} 
\begin{figure}[!h]
    \centering
    \includegraphics[width=\linewidth]{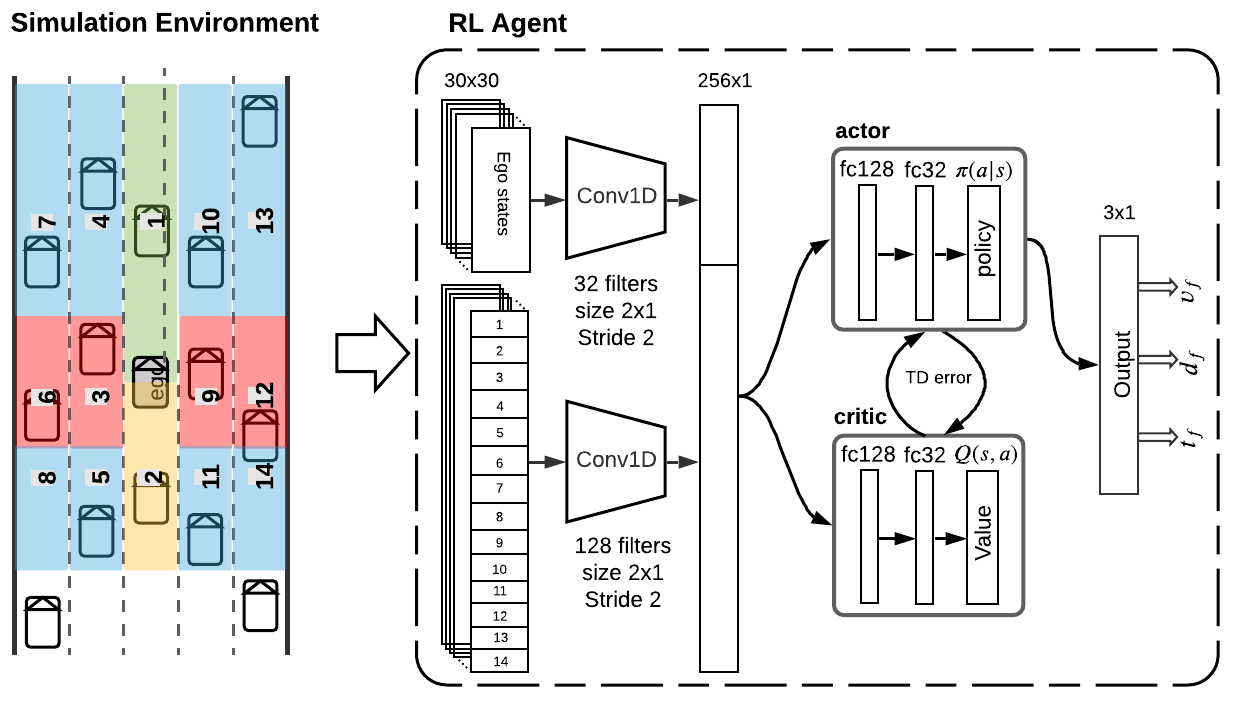}
    \caption{Neural network architecture of end-to-end continuous RL; 1D convolution layers encode the time-series data from vehicles' past trajectories.}
    \label{f:veh_config_state_repr}
\end{figure}

The observation space encloses the ego vehicle's states and the states of surrounding objects. We represent the input states $s$ to the neural network in Frenet frame (longitudinal $s$ and lateral $d$ displacements), where we normalize and convert them to $\xi$ before feeding them to the network i.e. $\xi_* \in [-1, 1]$. We choose egocentric state representation approach, where the positions and velocities of surrounding objects are expressed relative to the ego vehicle. Longitudinal and lateral displacements in Frenet frame, $s$ and $d$, describe both ego vehicle and surrounding objects states. The input states are $\xi_1$ and $\xi_2$ for the ego vehicle, and $\xi_{2i+1}$, and $\xi_{2i+2}$ for $i = \{1, 2, ..., N\}$ for the surrounding vehicles, where $N$ is defined as the maximum number of surrounding vehicles in the traffic. The input states are defined as in equation \ref{eq:norm_states}.
\begin{align}
\label{eq:norm_states}
    & \xi_1 = \frac{s_{ego} - s_{0}}{TL} \nonumber & \xi_{2i + 1} = s_{i} - s_{ego} \nonumber\\
    & \xi_2 = \frac{d_{ego}}{2 \, LW} \nonumber & \xi_{2i + 2} = d_{i} - d_{ego} \\
\end{align}
\noindent where longitudinal position of the ego $\xi_1$ is normalized w.r.t. its initial position and the total track length $TL$, and $2 \times LW$ (lane width) is used to normalize the lateral position state of the ego. $N=14$ discretized regions are defined around the ego vehicle, current lane front and back, immediate left/right, left/right up and left/right down of one and two adjacent lanes, which are shown in Fig. \ref{f:veh_config_state_repr}. We construct the input tensor for the neural network model based on the normalized states of the vehicles in those regions computed by equation \ref{eq:norm_states}. If a region is unoccupied, the corresponding value in the input tensor get $-1$. 

Since we are dealing with a dynamic system that evolves through time, we use a history of the actors' past trajectories. The input tensor is of shape $30\times30$, which encodes $30$ states of actors for past $30$ time steps. The states of all vehicles are stacked to express their evolution through time as illustrated in Fig. \ref{f:veh_config_state_repr}. We use convolution operation along time channel to extract the encoded time-series features from the past trajectories.
\subsubsection*{Neural Network Design} 
A neural network in DRL estimates the prior probabilities of taking different actions from action space $\mathcal{A}$ and the value of the current state in state space $\mathcal{S}$. We use two different 1D convolution layers separately for ego and surrounding vehicles' states as the feature extraction backbone, since they are defined in separate reference frames; ego states are represented w.r.t a fixed reference point (inertial Frenet frame), but surrounding vehicles' are relative to the ego vehicle (body Frenet frame). We then concatenate the encoded features to learn a policy upon them.
\subsection{Reinforcement Learning}

Reinforcement Learning algorithms mainly use Q-learning and/or policy gradient optimization for learning the control policy whether discrete or continuous control is needed. 

In Q-learning, the agent learns tries to learn the optimal state-value function $Q^*$ defined as in equation \ref{eq:Q-function}.

\begin{equation} \label{eq:Q-function}
    Q^*(s, a) = max_{\pi} \mathrm{}{E}(R_t| s_t=s, a_t=a, \pi)
\end{equation}

The action-state value function follows the Bellman equation,

\begin{equation}
    Q^{∗}(s, a) = \mathop{{}\mathbb{E}}[ r + \gamma max_{a^{'}} Q^{∗} (s^{'}, a^{'})|s, a]
\end{equation}

\noindent which is based on the intuition that if the values of $Q^{*}(s^{'}, a^{'})$ are known, the optimal policy is to select an action $a^{'}$, that maximizes the expected value of $Q^{*}(s^{'}, a^{'})$.

As the dimension of states and actions become large, solving q-learning by value or policy iteration becomes infeasible. The Deep Q-Network (DQN) algorithm uses a neural network with weights $\theta$ to approximate the optimal action-value function as $Q(s, a; \theta) \approx Q^{*}(s, a)$. Since the action-value function follows the Bellman equation, the weights can be optimized by minimizing the loss function as in equation \ref{eq:DQN_loss}.

\begin{equation} \label{eq:DQN_loss}
    L(\theta) = \mathop{{}\mathbb{E}}_{M}[(r + \gamma max_{a^{'}} Q(s^{'}, a^{'}; \theta^{-}) − Q(s, a; \theta))^{2}]
\end{equation}

\noindent where $M$ is mini-batch and $\theta^{-}$ the target neural network weights.

Continuous control reinforcement learning use policy gradient optimization technique to directly learn a policy over the state space, which is more efficient than first learning the state-action value and use it to derive the policy. The algorithms using policy gradient optimization try to maximize the objective function in equation \ref{eq:policy_gradient_optimizaiton} via gradient ascent.

\begin{equation} \label{eq:policy_gradient_optimizaiton}
    \nabla_{\theta} J(\theta) = \mathop{{}\mathbb{E}}_{\tau \sim \pi_{\theta}(\tau)} \bigg[ \bigg(\sum_{t=1}^{T} \nabla_{\theta} log \pi_{\theta}(a_t | s_t) \bigg) \bigg( \sum_{t=1}^{T} r(s_t, a_t) \bigg) \bigg]
\end{equation}

\noindent where $\tau$ is a trajectory, $\pi_{\theta}$ is the likelihood of executing a trajectory using $\pi_{theta}$ policy, $\pi_{theta}(a_t|s_t)$ the likelihood of executing action $a_t$ at state $s_t$, and $r(s_t, a_t)$ the returned reward gained for executing $a_t$ at state $s_t$.

The advantages of both aforementioned value-based and policy-based algorithms are merged into Actor-Critic method, where it takes the faster convergence property of the policy-based methods and the sample-efficiency of value-based methods. The actor model outputs the best actions through learning the optimal policy, and the critic model on the other hand evaluates the actions by computing their value functions. For the continuous control applications, the actor and critic models are parameterized mostly with neural network. We use the actor critic style of four reinforcement learning algorithms, named Proximal Policy Optimization (PPO) \cite{schulman2017proximal}, Trust Region Policy Optimization (TRPO) \cite{schulman2015trust}, Deep Deterministic Policy Gradient (DDPG) \cite{lillicrap2015continuous}, and Advantage Actor Critic (A2C) \cite{mnih2016asynchronous} as the successful widely-used reinforcement learning methods.




\section{EXPERIMENTS}
This section provides a comprehensive evaluation and performance comparison between the presented agents, i.e., baseline optimization, discrete RL and optimization, and continuous end-to-end RL. Since the baseline agent's driving style highly depends on the MOBIL, IDM, and the optimization parameter tuning, we have provided two sets of configurations: agile baseline and safe baseline. As the names suggest, each agent offers a different approach in the speed-maximization-lane-change-minimization trade-off. We utilized a simple trial and error method to tune the objective function coefficients. However, parameters that characterize hard constraints are selected according to the vehicle dynamics ($a_{max}$), highway regulations ($v_{max}$), and the safety criteria ($r$ and $v_{min}$) that we defined.

\subsection{Simulation Environment}
We have leveraged the CARLA v0.9.9.2 \cite{Dosovitskiy17} as the high-fidelity urban driving platform. The simulation presents a plausible alternative for actual driving scenarios in vehicle dynamics, environment stochasticity, and sensory measurements. We selected the Tesla Model 3 as the ego vehicle. We also utilized random locations in the highway loop available in Town04 to generate traffic scenarios.

\subsection{Training Process}
Since RL observations do not include any information regarding the road features, such as the curvature, there would be no point in using different highway maps between training and evaluation scenarios. However, each episode starts in a different location of the track. To generate the highway traffic for each episode, we created a list of vehicles with various dynamics available in CARLA and randomly spawned them around the ego. Each vehicle is equipped with an IDM and a range sensor for the speed control and does not perform lane change. The initial locations and the target speeds are randomly sampled from feasible ranges for each vehicle. To prepare the RL observation vector, we have used the ground truth information and appended a Gaussian noise to mimic the environment modeling error using 2D/3D object-detection algorithms.

\subsubsection*{Reward function} Each episode starts in a random location on the highway with random traffic surrounding the ego. The episode is terminated if ego drives for 500 (m), goes off the road or makes a collision. The course-grained reward function is defined as follows,
\begin{equation} \label{eq:coarse_gained_reward}
    r(s, a)= 
\begin{cases}
    -10,             & \text{if  collision}\\ 
    -10,              & \text{if goes off the road}
\end{cases}
\end{equation}
which motivates both the speed maximization and the passenger comfort. We define a fined-grained reward function which complements the equation \ref{eq:coarse_gained_reward}, 
    \begin{align}
        r_v &= w_{v}^{+} exp( \frac{ - err_v ^{2} }{ w_{v} * v_{max} } ) \label{eq:rew_v} \\
        r_{lc} &= \begin{cases} 
            r_v + r_v * w_{lc^+}, & \text{if speed gain $\%$ $>$ threshold} \\
            -1 * r_v * w_{lc^-},                     & \text{otherwise} 
        \end{cases} \label{eq:rew_lc} \\
        r(s, t) &= r_{v} + r_{lc} \label{eq:fine_gained_reward}
    \end{align}
\noindent where $r$ is the reward function, $v$ and $lc$ the speed and lane change. As defined in equation \ref{eq:rew_lc}, lane change can contribute both positive and negative in the reward function, whether the lane change results in speed gain or not. We define the threshold of speed gain as $0.08$ causing the positive lane change reward if lane change results in speed gain of at least $1$ m/s, otherwise the lane change maneuver gets a penalty. $w_{lc^+}$ and $w_{lc^-}$ are defined as $.07$ and $0.2$ respectively. The $w_{v^{+}}=10$ and $w_{v}=5$ are speed reward magnitude and shaping parameter of speed reward (defines speed reward bell-shape).

\begin{figure}[!h]
    \centering
    \includegraphics[width=\linewidth]{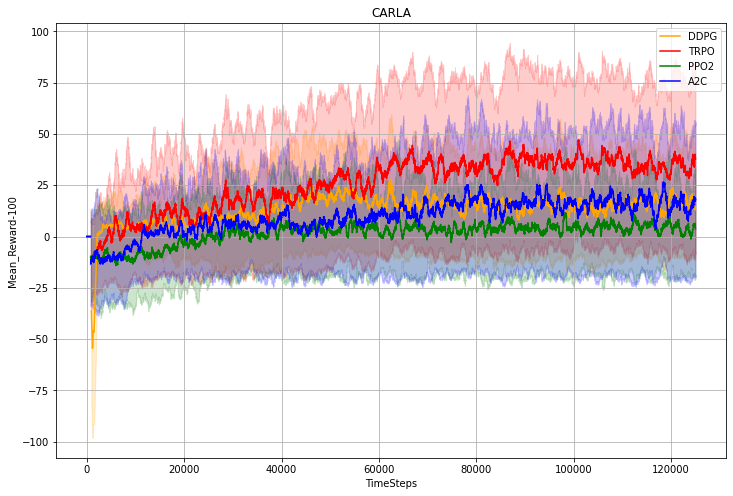}
    \caption{Average cumulative episode reward of RL agents in the training process}
    \label{fig:training_graph}
\end{figure}

Average cumulative episode rewards with window size 100 are presented at Fig. \ref{fig:training_graph}. TRPO achieves the highest cumulative reward, however it has the highest standart deviation. DDPG and A2C converges at a similar reward, but A2C also has higher standart deviation. During the training, PPO2 failed to converge at a reasonable policy. Overall, TRPO accumulated the highest episode reward at convergence, while DDPG has the lowest standart deviation due to the its memory. 

\subsection{Qualitative and Quantitative Analyses}
It is possible to compare the agents based on the RL's reward function (or the optimization cost function) that summarize the expected performance. However, this approach would undermine the credibility of the evaluation since each agent is expected to outperform the rivals in terms of its own reward function.  Alternatively, we provide both a qualitative and quantitative analyses of the agents' performance. The qualitative analysis has been conducted in several case studies. The scenarios cover situations where a self-driving car may commonly face in highway driving task, in addition, they target the intelligence, maneuverability, and the robustness of the agents. Quantitative analysis, on the other hand, has been performed on 1000 randomly generated traffic scenarios that were consistent for all agents. The following percentage metrics have been used to compare the agents' performance on all of the scenarios,
\begin{itemize}
    \item Speed = $100(1 - \frac{\vert \text{average speed} - \text{target speed}\vert}{\text{max error}})$
    \item Comfort = $100 (1-w_c\frac{\text{average jerk}}{\text{max jerk}} - (1-w_c)\frac{\text{average yaw rate}}{\text{max yaw rate}})$
    \item Safety = $\frac{100}{n}\sum_{i=1}^{n} (1 - \frac{\text{min TTC}}{TTC_i})$
\end{itemize}
where $w_c$ is used to weigh the importance of factors and TTC stands for the frontal time-to-collision experiences. In the Safety equation we included $n$ steps of the episode where ego tailgates a vehicle (TTC exists).

\subsubsection*{Case Study 1 (Intelligence)}
\begin{figure}[!h]
    \centering
    \includegraphics[width=\linewidth]{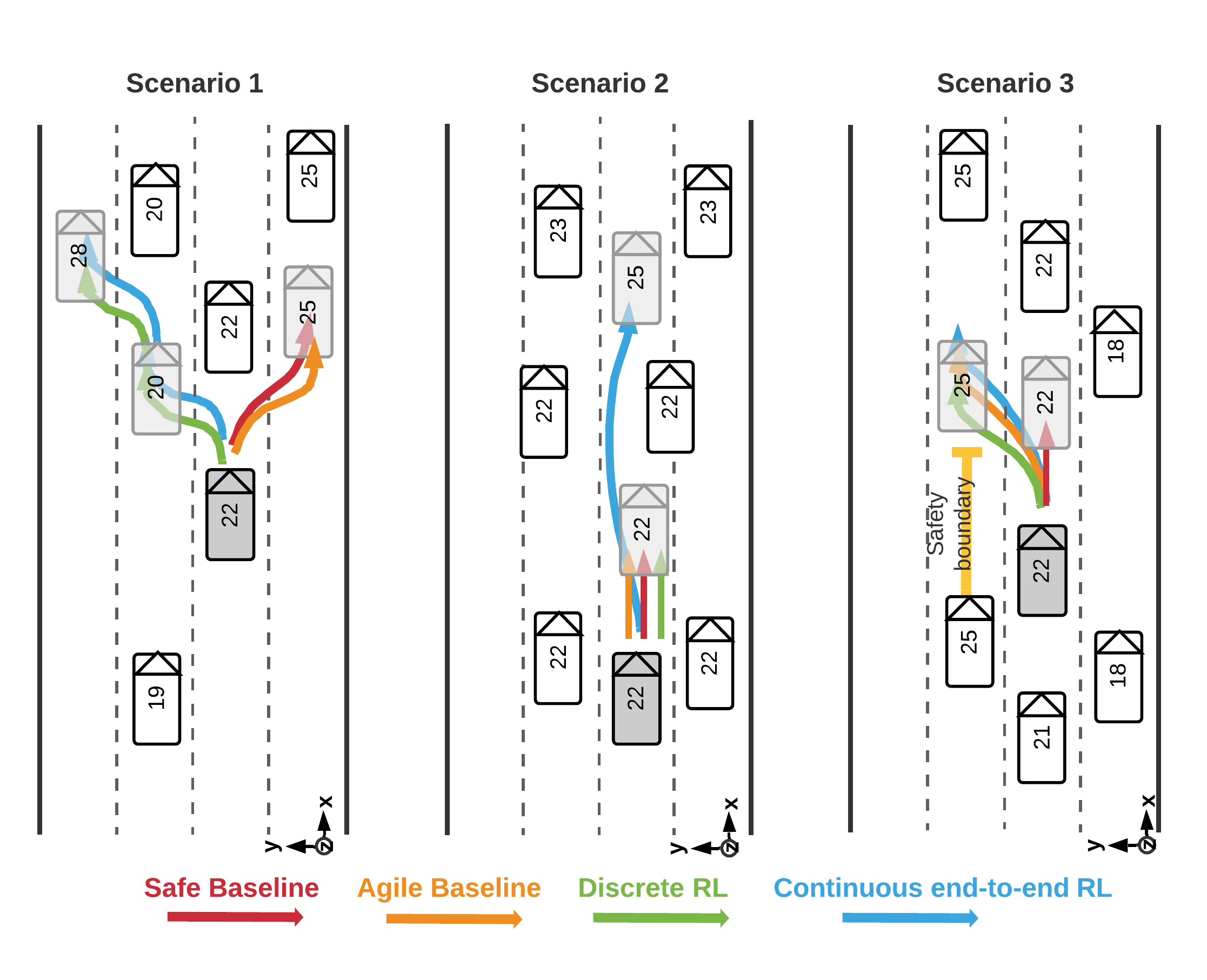}
    \caption{ Qualitative results for four proposed agents in three case studies. Arrows show the agents generated trajectory. Vehicles velocities are provided in m/s inside each vehicle's rectangle. Transparent ego shows the updated ego state after following the generated path. Videos recorded of the agents' performance for each scenario on CARLA are provided as supplementary files during the submission.}
    \label{fig:case studies}
\end{figure}

The scenario starts with the situation where the ego is driving in a middle lane. The left lane's traffic moves slower than the ego lane, contrasting with the right lane, which is faster than the ego lane. The far left lane, though, is empty. As illustrated in Fig.  \ref{fig:case studies}, both baseline agents change lane to the faster adjacent lane. This decision results in a faster ego and lower arrival time. On the other hand, RL agents move to the left lane before changing lane to the empty lane on the far left side. Although this action temporarily reduces ego speed, RL agents achieve higher average velocity throughout the scenario, helping them arrive at the destination (500m away) earlier than baselines. Besides, End-to-End RL agent gets back to the original lane after overtaking all actors. This strategic action enables the agent to have more freedom in upcoming traffic situations. Table \ref{tab:case studies} summarizes the quantitative percentages that each agent obtained. This scenario demonstrates the superiority of the AI agents in searching for globally optimal policies in complex traffic circumstances, resulting in a more intelligent driver than baselines. RL agents are well-known to maximize the accumulated future rewards, even though optimal policy provides cheap immediate rewards. In contrast with baselines that attempt to optimize the cost function without considering the future traffic situations.
\begin{table}
    \centering
      \begin{tabular}{c | c c c c}
         & Safe & Agile & Discrete & End-to-End\\
         & Baseline & Baseline & RL & RL\\
         \hline
         Speed & 76, 70, 63 & 80, 70, 63 & 93, 72, \textbf{72} & \textbf{94}, \textbf{87}, 71\\
        \hline
        Safety & \textbf{60}, \textbf{65}, \textbf{64} & 33, 61, 23 & 48, 62, 25 & 54, 36, 27\\
        \hline
        Comfort & \textbf{90}, 97, \textbf{94} & 80, 93, 78 & 72, \textbf{98}, 78 & 76, 82, 80\\
      \end{tabular}
    \caption{Performance metrics for three case studies as percentages (scenario 1, 2, and 3)}
    \label{tab:case studies}
\end{table}




\subsubsection*{Case Study 2 (Maneuverability)}
Safely navigating through vehicles to escape dense traffic is not an easy task for the agents. In this scenario, the ego is tailgating two vehicles driving off-center lanes. All vehicles surrounding the ego are driving at a constant low speed. Fig. \ref{fig:case studies} shows that all of the discrete agents continue tailgating the vehicle, which results in a safe but slow policy. Since the lattices generated by the local planner terminate at the lane centers, none of the discrete agents were competent in creating the escape passage through the traffic. Continuous RL, however, created a safe and feasible trajectory through the vehicles assisting the agent to avoid traffic and gain speed subsequently. This example verifies that continuous RL explores the driving corridors (Fig. \ref{f:driving corridor}) more carefully than other agents which presents a better maneuverability. Although reducing the lateral discretization ($d_f$) step-size in discrete agents would encourage the agents to create a similar passage, this approach would penalize the computation. Overall, this scenario shows that discrete agents may suffer the performance-computation trade-off while the continuous agent has a fixed, and possibly smaller, computation cost than other agents. The same circumstance holds for the lateral states' discretization ($v_f$ and $t_f$) and can be shown in similar scenarios, but we avoid doing this to save space in the paper.

\subsubsection*{Case Study 3 (Manipulation)}
This scenario depicts a typical situation for a vehicle that may face in a highway driving task. The ego is driving in a slower lane than the adjacent lane. A merging scenario would take ego to the faster lane. However, the safety of the merging scenario depends on the speed difference between the two lanes. Depending on the driver's style, a conservative driver may choose to stay in the current lane and drive safely. In contrast, an aggressive driver may overlook the merging risk and change lane to gain speed. We manually manipulated the baseline agents' parameters such that Safe Baseline stays on the current lane while Agile Baseline merges to the faster lane, even though the safety boundary of the following vehicle in the target lane is close. Likewise, both RL agents decided to neglect the safety risk and perform the merging action. This scenario shows the possibility of manipulating the baseline agents to apply a specific driving style. The same approach for the RL agents requires reshaping the reward function, which is more costly by a significant factor.

\subsubsection*{Quantitative Analysis}
In the qualitative analyses, we showed that baseline performance is highly dependent on parameter tuning and the information incorporated in the objective function. RL agents are more intelligent than baselines since they extract and leverage meaningful information directly from raw measurements. Here we compare the agents' performance on randomly generated scenarios. The vehicles are spawned in random relative positions and target speeds to the ego. The scenarios start in an arbitrary position on the highway track, and the track length is 500 meters. Surrounding vehicles are randomly selected from a vehicle pool in CARLA, each one having different dynamics. We evaluated the agents' performance in 1000 scenarios and recorded the results in Table \ref{tab:random scenarios}. Overall, the Safe Baseline showed better results in terms of safety and comfort. However, this came with the price of penalizing the speed with a notable factor - making the agent slowest, among others. A similar situation holds for Agile Baseline, where speed is preferred over safety and comfort. Although Baseline agents outperformed in terms of individual performance metrics, the End-to-End RL agent demonstrated a reasonable compromise, leading the agent to defeat others on average. This comparison demonstrates how RL agents can target multiple goals and attain a precise balance point in between.

\begin{table}
    \centering
      \begin{tabular}{c | c c c c}
         & Safe & Agile & Discrete & End-to-End\\
         & Baseline & Baseline & RL & RL\\
         \hline
         Speed
         & 55 \DrawBlackPercentageBar{0.55} & 81 \DrawRedPercentageBar{0.81} & 74 \DrawBlackPercentageBar{0.74} & 76 \DrawBlackPercentageBar{0.76}\\
        \hline
        Safety
        & 58 \DrawRedPercentageBar{0.58} & 28 \DrawBlackPercentageBar{0.28} & 41 \DrawBlackPercentageBar{0.41} & 52 \DrawBlackPercentageBar{0.52}\\
        \hline
        Comfort
        & 81 \DrawRedPercentageBar{0.81} & 47 \DrawBlackPercentageBar{0.47} & 70 \DrawBlackPercentageBar{0.70} & 78 \DrawBlackPercentageBar{0.78}\\
        \hline
        Average & 65 \DrawBlackPercentageBar{0.65} & 52 \DrawBlackPercentageBar{0.52} & 62 \DrawBlackPercentageBar{0.62} & 69 \DrawRedPercentageBar{0.69}\\
      \end{tabular}
    \caption{Performance metrics for 1000 randomly generated test scenarios as percentages}
    \label{tab:random scenarios}
\end{table}


\section{CONCLUSIONS}
The results of this paper show that the presented method, which combines the decision making and motion planning in an end-to-end manner, can create agents capable of making decisions for long-horizon and optimally generating local trajectories. For three different highway driving cases, the end-to-end continuous deep RL performs the best amongst baselines and discrete RL agent. As an example, for generating inter-lane trajectories, the continuous RL (presented method) performs optimally while maintaining the computational complexity, but its competitors require more computation to address the same problem.

\bibliographystyle{IEEEtran}
\bibliography{references.bib}

\vspace{12pt}
\color{red}

\end{document}